\documentclass[runningheads]{llncs}
\usepackage{eccv}

\usepackage{eccvabbrv}

\usepackage{graphicx}
\usepackage{booktabs}
\usepackage[dvipsnames]{xcolor}
\usepackage[table,xcdraw]{xcolor}
\usepackage{arydshln}
\usepackage{ulem}
\usepackage{multirow}
\usepackage{overpic}

\usepackage[pagebackref,breaklinks,colorlinks,citecolor=eccvblue]{hyperref}

\usepackage[accsupp]{axessibility}

\usepackage{hyperref}

\usepackage{orcidlink}

\begin{document}

\title{DINO-SLAM: DINO-informed RGB-D SLAM for Neural Implicit and Explicit Representations}

\titlerunning{DINO-SLAM}

\author{
	Ziren Gong$^{1*}$\orcidlink{0000-0003-0093-835X} \quad\quad Xiaohan Li$^{1,2*}$\orcidlink{0000-0003-3109-9413} \quad\quad Fabio Tosi$^{1}$\orcidlink{0000-0002-6276-5282} \quad\quad Youmin Zhang$^{3}$\orcidlink{0009-0000-5409-9171} \\ Stefano Mattoccia$^{1}$\orcidlink{0000-0002-3681-7704} \quad\quad Jun Wu$^{4}$\orcidlink{0000-0001-7090-8653} \quad\quad Matteo Poggi$^{1}$\orcidlink{0000-0002-3337-2236} 
}

\authorrunning{Gong et al.}

\institute{University of Bologna, Italy \quad\quad\quad $^*$ joint first authorship \and
Faculty of Dentistry, The University of Hong Kong, China \and
Rawmantic AI, China \and Fudan University, Shanghai, China \\
\texttt{Project page:} \url{https://zorangong.github.io/DINO-SLAM/}
}

\maketitle
\begin{center}
    \vspace{-0.3cm}
    \begin{tabular}{l c c c}
        \includegraphics[width=\linewidth,scale=1.00]{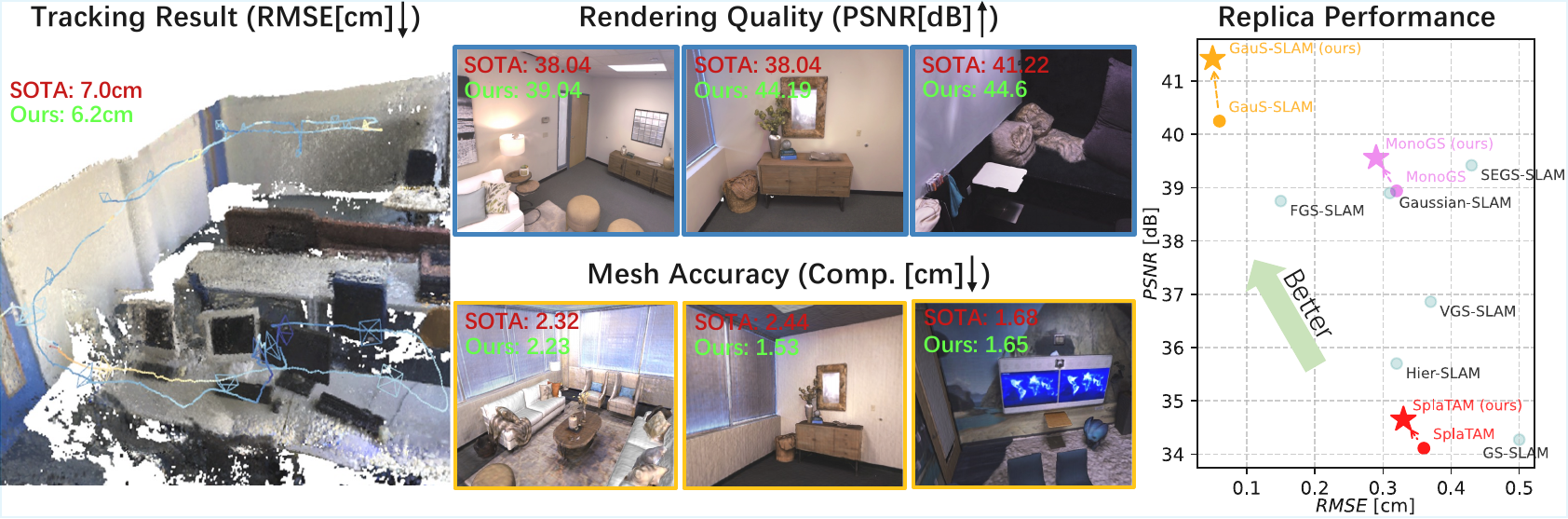}
    \end{tabular}
\end{center}\vspace{-0.7cm}
\captionof{figure}{\textbf{DINO-SLAM in action on Replica \cite{straub2019replica}.} By enhancing DINO features with geometry awareness, we design a DINO-informed SLAM pipeline achieving superior camera tracking accuracy (on the left, evaluated on \textit{ScanNet}) and 3D reconstruction quality (in the middle, evaluated on \textit{Replica}). Our approach is general: it can be integrated into existing NeRF/GS SLAM systems and boost their results (on the right, showing some representative GS methods being improved by our proposal).
}\label{fig:teaser}

\begin{abstract}
This paper presents DINO-SLAM, a DINO-informed design strategy to enhance implicit (Neural Radiance Field -- NeRF) and explicit representations (Gaussian Splatting -- GS) in SLAM systems through the more comprehensive semantic  understanding enabled by DINO.
This latter alone, however, lacks proper 3D geometry understanding, allowing only for marginal improvements.
Therefore, we rely on a Scene Geometry Encoder (SGE) to lift DINO features into geometry-aware DINO features (geoDINO), to better understand those geometric relationships that vanilla DINO features fail to capture. Building upon it, we propose two foundational paradigms for NeRF and GS SLAM systems integrating geoDINO features. 
Compared to state-of-the-art methods, our DINO-informed pipelines achieve superior performance on the Replica, ScanNet, and TUM datasets.
\end{abstract}

\section{Introduction}
\label{sec:introduction}
 
Dense simultaneous localization and mapping (SLAM) has emerged as a fundamental technology in robotics and autonomous systems, with extensive applications ranging from mobile robotics to autonomous vehicles. As a core component of spatial intelligence, dense SLAM enables real-time camera localization while constructing detailed environmental maps, thereby facilitating autonomous navigation and scene understanding. Initially, SLAM techniques were primarily hand-crafted algorithms \cite{orbslam, mur2017orb, campos2021orb, newcombe2011dtam, salas2013slam++}, facing limitations in the presence of challenging lighting conditions and poorly textured environments. The advent of deep learning has further advanced the field, significantly improving pose estimation accuracy through learned features and matching \cite{tateno2017cnn, li2020structure, teed2021droid}, peaking with the recent breakthroughs in novel view synthesis, notably Neural Radiance Fields (NeRF) \cite{mildenhall2021nerf} and Gaussian Splatting (GS) \cite{kerbl20233d}, that have revolutionized this field with new, exciting possibilities. Among these, the possibility of rendering images from novel viewpoints during or after the mapping process is an intriguing side-product not enabled by conventional SLAM systems or even by the most modern feed-forward models \cite{murai2025mast3r,liu2025slam3r,maggio2025vggt-slam,gong2026magist3r}. 
Nevertheless, current NeRF or GS-based SLAM systems primarily focus on geometric reconstruction and photometric consistency, overlooking higher-level scene understanding necessary for complex scene representation and autonomous navigation. 
Indeed, on the one hand, these frameworks rely on supervisory signals coming from pixel-wise colors or depth, thus lacking such a higher-level understanding of both those local and global properties that conventional feed-forward vision models can encode in deep features \cite{caron2021emerging,oquab2023dinov2}. 
On the other hand, vision foundation models like DINO \cite{caron2021emerging,oquab2023dinov2} excel at extracting high-level features from color images, yet inherently do not take into account the 3D structure of the scene. Therefore, we argue that naively integrating such features into SLAM pipelines could already soften the aforementioned problem, yet only marginally. By contrast, enriching such features with geometry awareness has the potential to further improve performance: we argue this can be achieved by jointly processing such features with the simplest geometry cue available to RGB-D SLAM systems: \textbf{depth}.

In this paper, we propose DINO-SLAM, a novel and general framework that can seamlessly integrate the higher-level scene understanding of a self-supervised foundation model like DINO \cite{caron2021emerging,oquab2023dinov2} into the SLAM system. This is achieved by embedding the features extracted by the foundation model directly into the scene representation, either implicit or explicit, learned during the optimization carried out by the SLAM system, yet only after having lifted such features to encode 3D geometry properties of the scene.
The foundation of DINO-SLAM is a custom model, the \textbf{Scene Geometry Encoder (SGE)}, that jointly processes the DINO features extracted from the color frame with its corresponding depth map to extract a hierarchy of new, higher-level features, namely \textbf{geoDINO} features, encoding both contextual understanding and scene geometry. Building upon this extractor, we establish two specialized paradigms suited for both NeRF-based and GS-based SLAM methods.  
We demonstrate the wide applicability of our approach through integration with various scene-encoding SLAM methods \cite{wang2023co, johari2023eslam, sandstrom2023point, keetha2024splatam, matsuki2024gaussian}, over standard benchmarks in the field \cite{straub2019replica,sturm2012benchmark,dai2017scannet}, where DINO-SLAM achieves state-of-the-art accuracy -- as shown in Fig. \ref{fig:teaser}. 

Our key contributions are:
\begin{itemize}

\item We introduce DINO-SLAM, a DINO-informed RGB-D SLAM design paradigm significantly boosting the performance of existing SLAM pipelines by embedding enhanced DINO features within scene representations.

\item Purposely, we develop a Scene Geometry Encoder (SGE) to enrich the scene representations captured by DINO with geometry awareness retrieved from depth maps. The SGE is optimized online alongside the SLAM pipeline. Based on it, we implement two foundational paradigms suitable for both NeRF-based and GS-based SLAM systems.

\item We validate our wide applicability to neural implicit and explicit SLAM methods by integrating our SGE into various scene representations, such as hash grids, triplanes, 3D Gaussians Splats, and 2D Gaussian Surfels. Experiments on popular datasets \cite{straub2019replica, sturm2012benchmark, dai2017scannet} demonstrate the superior performance of our DINO-SLAM pipelines.

\end{itemize}

\section{Related work}
Here, we briefly review the research related to our work. More detailed literature can be found in \cite{tosi2024nerfs}.

\textbf{Neural Implicit SLAM}. With the emergence of NeRF \cite{mildenhall2021nerf}, numerous works \cite{sucar2021imap, zhu2022nice, johari2023eslam, wang2023co, deng2024plgslam, hsslam, sandstrom2023point} have incorporated such implicit volumetric representation into SLAM frameworks.
As the first approach using implicit neural representations for SLAM, iMAP \cite{sucar2021imap} adopts an MLP to map the 3D coordinates to color and volume density. NICE-SLAM \cite{zhu2022nice} further extends the single MLP of iMAP into a hierarchy of MLPs, which improves the quality of dense reconstruction by optimizing a hierarchical representation.
Current works further explore different scene encodings, such as hash-tables \cite{wang2023co, zhang2023go, xin2024hero, park2024lrslam,snh-slam}, neural-point \cite{sandstrom2023point, liso2024loopy, hu2023cp}, and tri-planes \cite{johari2023eslam, deng2024plgslam, park2024lrslam}, to enhance scene modeling capabilities, or orthogonal features such as uncertainty \cite{mamba-slam} or quantized queries \cite{qq-slam}.

\textbf{Neural Explicit SLAM}. Gaussian Splatting (GS) \cite{kerbl20233d}, as an explicit radiance field technique, has been recently widely used in SLAM methods. Compared to NeRF-based SLAM, these methods \cite{keetha2024splatam, yan2024gs, yugay2023gaussian, matsuki2024gaussian, huang2024photo, hu2024cg, peng2024rtg, ha2024rgbd, zheng2025wildgs} offer two key advantages: explicit spatial extent mapping and local explicit mapping editability, enabling real-time adjustments and corrections of the scene representations. As pioneers, MonoGS \cite{matsuki2024gaussian} incorporates GS into RGB-D SLAM systems for efficient and high-quality rendering of environments. 
SplaTAM \cite{keetha2024splatam} leverages a group of simplified Gaussians to represent the whole scene, GS-SLAM \cite{yugay2023gaussian} encodes both appearance and geometry by 3D Gaussians with opacity and spherical harmonics, while WildGS-SLAM \cite{zheng2025wildgs} learns to ignore moving objects and distractors through uncertainty. Other GS-based systems focus on semantic SLAM \cite{hier-slam,vsg-slam}, adding appearance embeddings \cite{segs-slam} or opacity fields \cite{fgo-slam}, performing adaptive densification based on Fourier frequency domain analysis \cite{cgs-slam}, reducing and compressing Gaussians \cite{cgs-slam}, or
replacing 3D Gaussians with 2D Gaussian Surfels \cite{vss-slam,su2026gaus}.  
 
\textbf{DINO-informed Neural Representation}. DINO \cite{caron2021emerging} has emerged as a powerful tool in neural scene representations, offering both invariant feature extraction \cite{mazur2023feature, xu2022sinnerf} and semantic comprehension \cite{ye2023featurenerf, fan2022nerf, dou2024learning}. Mazur et al. \cite{mazur2023feature} extract DINO features that offer flexible semantic representations for scene understanding. 
Recent works \cite{zhu2024semgauss, zhu2024sni, singh2024loss, yu2022d, pham2024go} have already investigated the integration of DINO \cite{caron2021emerging} or DINOv2 \cite{oquab2023dinov2} features into SLAM pipelines, although for different purposes -- i.e., for joint semantic segmentation, mapping, and localization, or to identify and filter out moving subjects in dynamic SLAM pipelines. 

As a consequence, both implicit and explicit methods overlooked the potential of DINO features and how their guidance, if properly enriched with geometry awareness, can improve the accuracy of existing SLAM systems. In this paper, we comprehensively explore this path and unveil the real potential of DINO, when properly enhanced with geometry awareness, to boost a variety of SLAM systems characterized by different scene representations, thus suggesting the potential to extend our approach to future SLAM systems based on radiance fields. 

\begin{figure*}[t]
	\centering
    \begin{overpic}
        [width=\linewidth,scale=1.00]{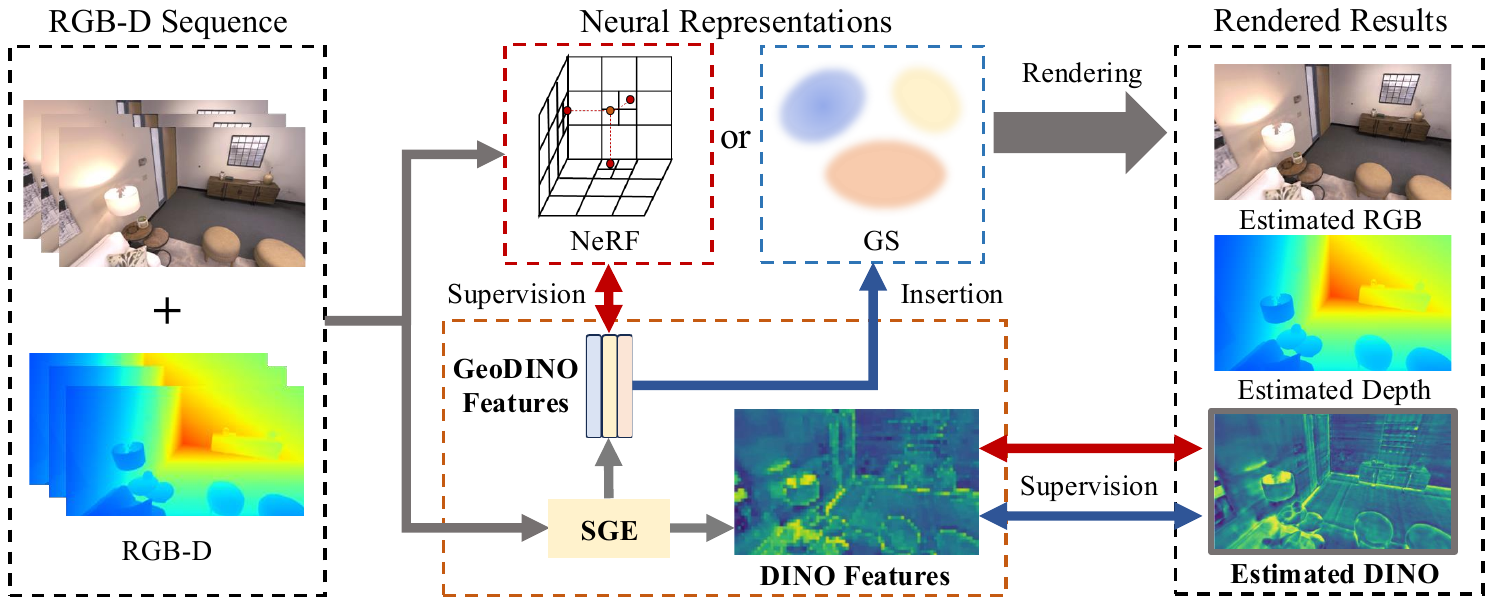}
    \end{overpic}\vspace{-0.3cm}
	\caption{\textbf{High-level overview of DINO-SLAM for neural implicit or explicit representations.} Both neural representations share the same structure of \textbf{Scene Geometry Encoder (SGE)} to capture geometry-aware DINO (geoDINO) and DINO features. In our neural implicit pipeline (NeRF), \textbf{geoDINO features} provide supervision for the tri-plane optimization (\textcolor{BrickRed}{\textbf{red double arrow}}), while \textbf{DINO features} serve roles in guiding the optimization of the estimated DINO feature map (\textcolor{BrickRed}{\textbf{red double arrow}}). In our neural explicit pipeline (GS), we incorporate \textbf{geoDINO features} into Gaussian parameters (\textcolor{RoyalBlue}{\textbf{blue arrow}}) and leverage \textbf{DINO features} to supervise the estimated feature map (\textcolor{RoyalBlue}{\textbf{blue double arrow}}).
 }
	\label{fig:overview}
\end{figure*}

\begin{figure*}[t]
	\centering
	\includegraphics[width=\linewidth,scale=1.00]{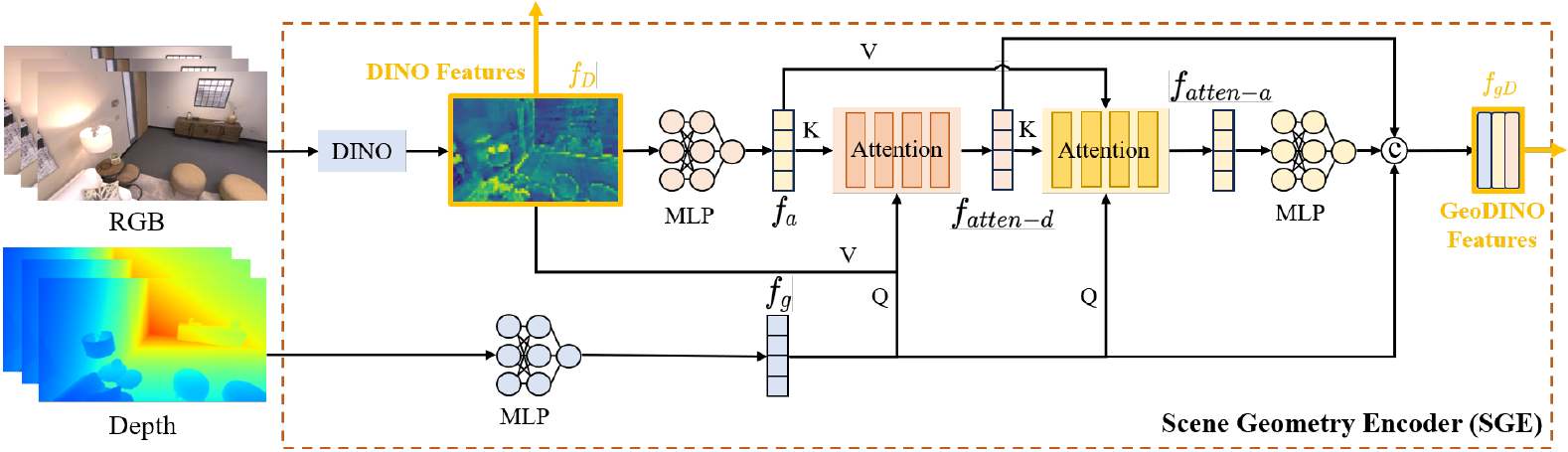}\vspace{-0.3cm}
	\caption{\textbf{Architecture of the Scene Geometry Encoder (SGE).} Our SGE has two outputs (see \textcolor{Goldenrod}{\textbf{yellow boxes}}): 1) \textbf{DINO features $(f_{D})$}, simply extracted from RGB frames by the frozen DINO model; 2) \textbf{geoDINO features $(f_{gD})$}, obtained from the joint processing of DINO features and depth, projected respectively into appearance and geometric features $f_{a}, f_{g}$, and used as keys and queries to attend over $f_{D}$.}
	\label{fig:DINO-extractor}
\end{figure*}

\section{Method Overview}

We now introduce our DINO-SLAM framework, which supports both neural implicit and explicit representations, as illustrated in Figure \ref{fig:overview}.

\subsection{Scene Geometry Encoder}
\label{subsec:DINO-extractor}

Given the strong priors encoded in DINO features \cite{caron2021emerging}, we design a Scene Geometry Encoder (SGE) to further enrich these embeddings with geometric understanding and generate more informative, geometry-aware DINO features (geoDINO). These cues guide the scene modeling in our SLAM systems, removing bottlenecks caused by conventional color and depth pixel-wise supervisions, as well as the lack of geometric knowledge in vanilla DINO features.
As shown in Figure \ref{fig:DINO-extractor}, our encoder can be formalized as a model $\Phi_{SGE}$ that processes the color $c$ and depth $d$ to extract both vanilla DINO $f_{D}$ and geometry-aware geoDINO features $f_{gD}$:
\begin{equation}
    \Phi _{SGE}\left ( c, d \right ) \to \left ( f_{D} ,f_{gD}  \right ) 
\end{equation}

Specifically, DINO features $f_{D}$ are obtained through the frozen, ViT-small DINOv2 model \cite{oquab2023dinov2} and are further processed with an MLP, projecting them into features $f_{a}$. Concurrently, depth frames are projected by a second MLP into geometric features $f_{g}$, to encode spatial relations and arrangements. Then, $f_{a}$ and $f_{g}$ are used as keys and queries to inject structural geometry information into features $f_{D}$ by means of an attention module, producing $f_{atten-d}$:
\begin{equation}
   f_{atten-d} =softmax\left ( \frac{f_{g}  f_{a}^{T} }{\sqrt{d_{a}} }  \right ) f_{D} 
\end{equation}
As a result, SGE forces DINO features to be re-weighted based on the local 3D structure, encoded by $f_{g}$, in relation to the whole depth map.
This mechanism is repeated, with $f_{atten-d}$ and $f_{g}$ being then used as keys and queries for attending over the previous projected features $f_{a}$ and obtaining enhanced features $f_{atten-a}$:
\begin{equation}
    f_{atten-a} =softmax\left ( \frac{f_{g}  f_{atten-d}^{T} }{\sqrt{d_{atten-d}} }  \right ) f_{a}
\end{equation}
Finally, a further projection performed through a third MLP produces $f_{atten-a}^{'}$, which are concatenated with $f_{atten-d}, f_{g}$ to obtain our geoDINO features. These features can be embedded inside any NeRF or GS-based SLAM systems: when doing so, our SGE is optimized online alongside the NeRF or GS backbone in the pipeline, by exploiting two feature losses detailed in the remainder.
Our experiments will demonstrate that these higher-level, geometry-aware features are crucial for achieving the best results, whereas embedding DINO features alone in SLAM pipelines yields marginal improvements.

\begin{figure*}[t]
	\centering
	\includegraphics[width=\linewidth,scale=1]{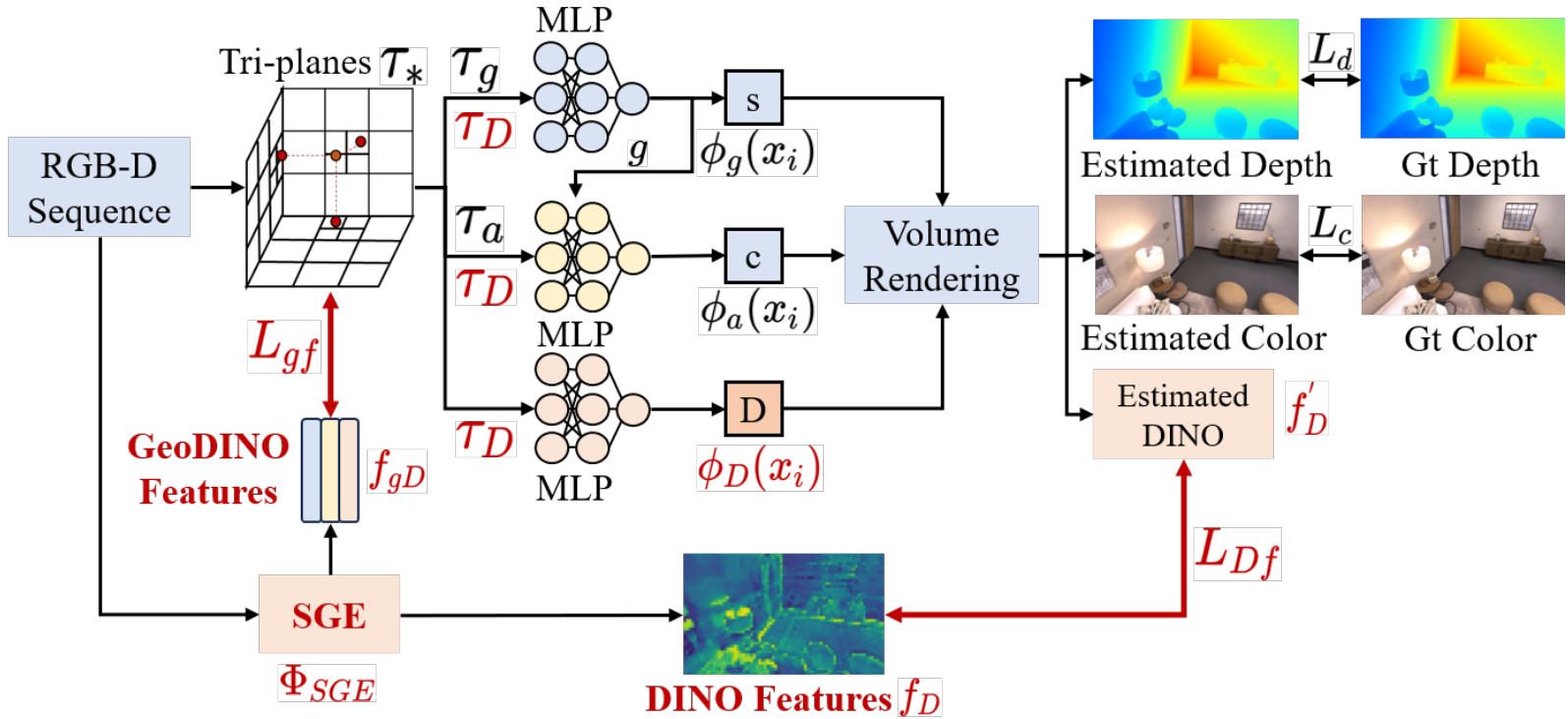}\vspace{-0.3cm}
	\caption{\textbf{Architectures of DINO-informed, NeRF-based SLAM 
 pipelines}. We leverage geoDINO features $f_{gD}$ to supervise features -- e.g., tri-plane $\tau _{D}$, while DINO features $f_{D}$ guide the optimization of the DINO feature map $f_{D}^{'}$ estimated by NeRF. 
 }\label{fig:nerf-pipelines}
\end{figure*}

\subsection{NeRF-based Pipeline}
\label{subsec:NeRF-pipline}

Building upon the features extracted by our SGE, we design a first NeRF-based version of DINO-SLAM, illustrated in Figure \ref{fig:nerf-pipelines}. 

\textbf{Scene Representation}. Our NeRF-based pipeline is optimized to render RGB, depth, and vanilla DINO features through ray sampling, leveraging the per-pixel ray casting property inherent to NeRF methods. As illustrated in Figure \ref{fig:nerf-pipelines}, our framework encodes the sampled points $x_{i}$ along the rays into some feature representation. In the figure, we show an implementation using tri-planes \cite{johari2023eslam} for which, specifically, we design three distinct tri-planes ($\tau_{g}$, $\tau_{a}$, and $\tau_{D}$) to learn geometric, appearance, and geoDINO representations separately. Then, three shallow MLPs, each consisting of two fully-connected layers, process these tri-plane encodings to produce the signed distance function (SDF) $\phi _{g}(x_{i}) $, raw colors $\phi _{a}(x_{i})$, and raw DINO features $\phi_{D}(x_{i})$.
While the vanilla DINO features extracted from images supervise $\phi_{D}$, the geoDINO features from our SGE supervise the tri-plane encodings $\tau_{D}$, as detailed in the subsequent sections.

\textbf{DINO-informed Rendering}. We adopt the standard ray-casting process used in NeRF \cite{mildenhall2021nerf}. For all points sampled along a ray $\left \{ p_{n}  \right \}_{n=1}^{N}$, we query TSDF $\phi _{g}(p_{n})$, raw color $\phi _{a}(p_{n})$, and raw DINO features $\phi _{D}(p_{n})$ from the MLPs. We then apply SDF-based rendering to compute volume densities:
\begin{equation}
    \sigma \left ( p_{n}  \right ) = \beta \cdot Sigmoid\left ( -\beta \cdot \phi _{g} \left ( p_{n} \right )  \right )
\end{equation}
where $\sigma \left ( p_{n}  \right )$ represents the volume density and $\beta$ is a learnable parameter controlling the sharpness of the surface boundary. The termination probability $w_{n}$, color $c$, depth $d$, and estimated DINO features $f_{D}^{'}$ are rendered using these volume densities:
\begin{gather}
    \quad f_{D}^{'}  = \sum_{n=1}^{N} w_{n} \phi _{D}\left ( p_{n}  \right ) \quad 
    \text{with} \quad w_{n} = exp\left ( -\sum_{k=1}^{n-1} \sigma \left ( p_{k}  \right )  \right )\left ( 1-exp\left ( -\sigma \left ( p_{n}  \right )  \right )  \right ), \notag   \\
    \quad\quad c  = \sum_{n=1}^{N} w_{n} \phi _{a}\left ( p_{n}  \right ),\quad \text{and} \quad
    d  = \sum_{n=1}^{N} w_{n} z_{n} 
\end{gather}
where $z_{n}$ denotes the depth of sampled point $p_{n}$.

\textbf{DINO-informed Supervision}. We introduce two feature-level losses to enhance scene representation and semantic consistency. As shown in Figure \ref{fig:nerf-pipelines}, we implement a geoDINO feature loss $L_{gf}$ to supervise the tri-planes $\tau _{D}$ responsible for learning enriched spatial and appearance cues. This loss function computes the feature similarity between the tri-planes $\tau _{D}$ and the geoDINO features $f_{gD}$:
\begin{equation}
    L_{gf} = \left \| \tau _{D} - f_{gD}  \right \| _{1} 
\end{equation}
We further introduce a loss $L_{Df}$ between predicted and vanilla DINO features: 
\begin{equation}
    L_{Df} = \left \| f_{D}^{'} - f_{D}  \right \| _{1}
\end{equation}
Additionally, we employ standard color and depth losses:
\begin{equation}
    L_{c}=\frac{1}{\vert \ R \ \vert}\sum_{r \in R}\left ( c_{n} - \hat{c}_{n}  \right )^{2}, \quad L_{d}=\frac{1}{\vert \ R \ \vert}\sum_{r \in R}\left ( d_{n} - \hat{d}_{n}  \right )^{2}  
\end{equation}
where $R$ denotes a batch of rays. $\hat{c}_{n}$ and $\hat{d}_{n}$ represent the ground-truth color and depth, respectively. We also incorporate additional loss terms from \cite{johari2023eslam}, including free space loss, signed distance loss, and truncation loss.

\textbf{Mapping and Tracking}. Following \cite{johari2023eslam}, our mapping process randomly samples $R$ pixels from a temporal window comprising the current frame, the two most recent frames, and $W-3$ frames selected from keyframes. Keyframes are systematically selected every $k$ frames from the input RGB-D sequences. The training process jointly optimizes three components: the tri-planes ($\tau_{g}$, $\tau_{a}$, and $\tau_{D}$), the decoder MLPs, and camera poses for the selected $W$ frames. For tracking, we perform pose estimation for every incoming frame.

\begin{figure*}[t]
	\centering
	\includegraphics[width=\linewidth,scale=1]{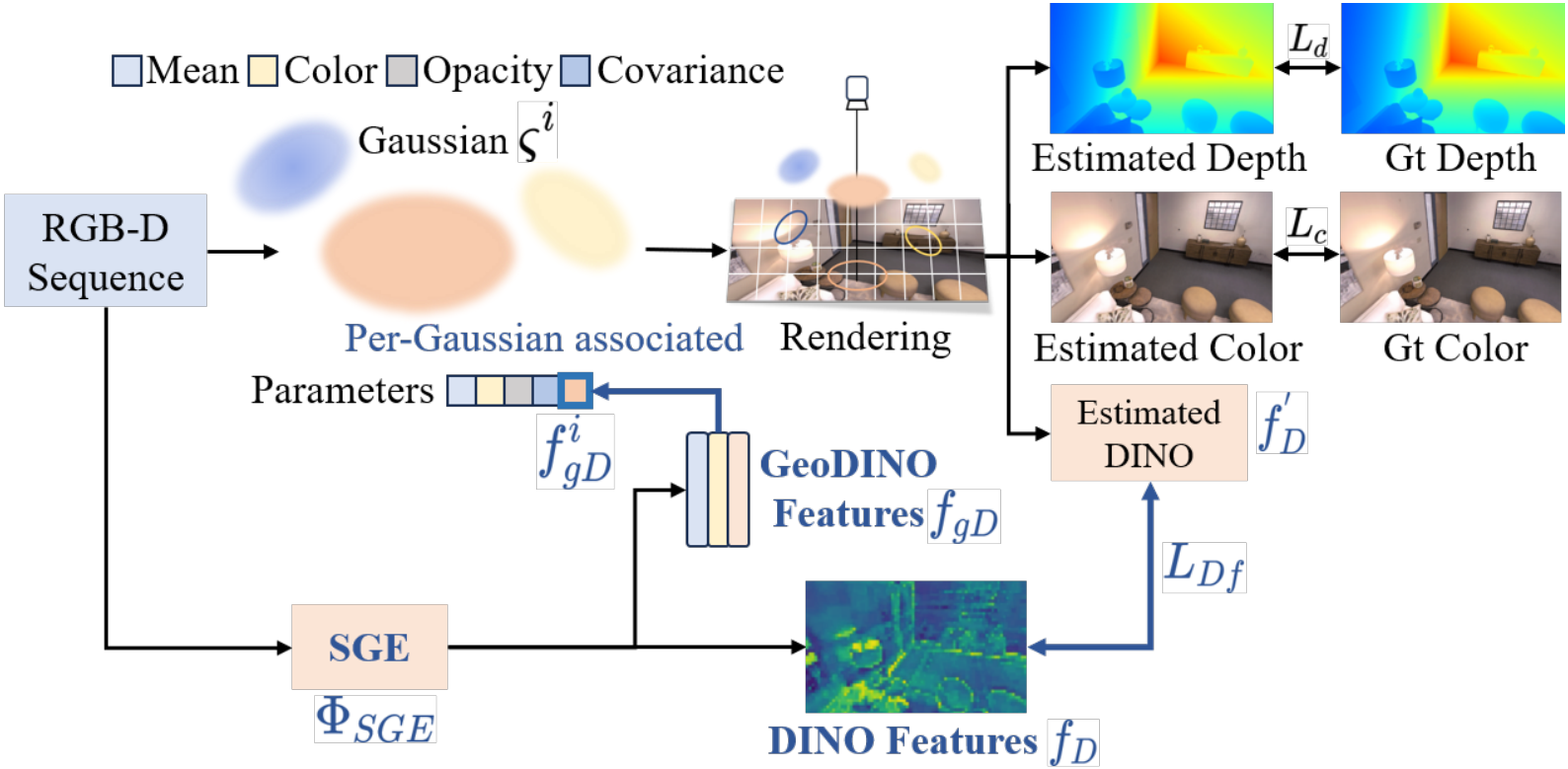}\vspace{-0.3cm}
	\caption{\textbf{Architectures of DINO-informed, GS-based SLAM 
 pipelines}. We incorporate the corresponding geoDINO feature $f_{gD}^{i}$ into the parameters of each Gaussian $\varsigma ^{i}$. Meanwhile, the DINO features $f_{D}$ guide the learning within the Gaussians.
 }
	\label{fig:3dgs-pipelines}
\end{figure*}

\subsection{GS-based Pipeline}

Driven by the same principles, we propose a second version of DINO-SLAM built over GS, shown in Figure \ref{fig:3dgs-pipelines}. 

\label{subsec:3DGS-pipline}
\textbf{Scene Representation}. Following vanilla GS, the scene is represented through a set of anisotropic Gaussians $\varsigma$ where each Gaussian primitive $\varsigma ^{i}$ is characterized by its parameters: color $c^{i}$, radius $r^{i}$, and opacity $o ^{i}$. For continuous spatial representation, each Gaussian is further defined by its mean $\mu ^{i}$ and covariance $\Sigma ^{i} $, representing its position and ellipsoidal shape, respectively.
As shown in Figure \ref{fig:3dgs-pipelines}, we incorporate the geoDINO features $f_{gD}$ obtained from our SGE into each Gaussian's set of parameters, while the vanilla DINO features extracted from images supervise the Gaussians themselves, emulating the NeRF-based pipeline introduced in the previous section. 
Specifically, each Gaussian $\varsigma ^{i}$ corresponds to an embedding in the geoDINO feature map. We assign its associated feature $f_{gD}^{i}$ into the distinct channel of Gaussian $\varsigma ^{i}$ parameters.
This enhanced representation can be formalized through the following Gaussian equation:
\begin{equation}\label{eq_varsigma}
    \varsigma \left ( x \right ) = o \exp\left ( -\frac{\left \| x-\mu  \right \|^{2}  }{2r^{2} }  \right )
\end{equation}

\textbf{DINO-informed Rasterization}.
Our approach employs iterative rasterization over Gaussians instead of per-pixel ray casting. During rasterization, we project the Gaussians onto the image plane to obtain 2D Gaussians through:
\begin{equation}\label{eq_mu}
    \mu_{2d}=\pi \left ( T_{CW} \cdot \mu  \right ),\Sigma _{2d}=JW\Sigma W^{T}J^{T}
\end{equation}
where $\mu_{2d}$ and $\Sigma _{2d}$ represent the position and the covariance of 2D Gaussians, respectively. $\pi$ denotes the projection operation and $T_{CW}$ is the camera pose of the viewpoint. $J$ is the Jacobian of the linear approximation of the projective transformation, and $W$ is the rotational component of $T_{CW}$. Through splatting and blending $N$ Gaussians, we can obtain the rendered feature map $f_{D}^{'}$ that captures enriched geometric and appearance information:
\begin{equation}
    f_{D}^{'} = \sum _{i\in N} f_{gD}^{i} \varsigma^{i} \prod_{j=1}^{i-1}(1-\varsigma ^{j} ) 
\end{equation}
Similarly, we obtain color $c$ and depth $d$ through:
\begin{equation}
    c = \sum _{i\in N} c^{i} \varsigma ^{i} \prod_{j=1}^{i-1}(1-\varsigma ^{j} ), \quad d = \sum _{i\in N} z^{i} \varsigma ^{i} \prod_{j=1}^{i-1}(1-\varsigma ^{j} )
\end{equation}
where $z^{i}$ is the distance to $\mu$ along the camera ray.

\textbf{DINO-informed Supervision}. We introduce a DINO-feature loss $L_{Df}$ to keep semantic consistency and guide the optimization of the enriched scene representation. As illustrated in Figure \ref{fig:3dgs-pipelines} (b), this loss function measures the feature similarity between the DINO features $f_{D}$ and the rendered feature maps $f_{D}^{'}$:
\begin{equation}
    L_{Df} = \left \| f_{D}^{'}-f_{D}  \right \| _{1} 
\end{equation}
Similarly, we define color loss and depth loss for the appearance and geometric optimization:
\begin{equation}
    L_{c} = \left \| c - \hat{c} \right \| _{1}, \quad L_{d} = \left \| d-\hat{d}  \right \| _{1}
\end{equation}
where $\hat{c}$ and $\hat{d}$ represent the ground-truth color and depth, respectively. 

\textbf{Mapping and Tracking}. For mapping and tracking, we adopt the keyframe selection strategy based on co-visibility from \cite{li2024ec}. The tracking thread focuses solely on camera pose optimization using color and depth residuals. In the mapping thread, depending on the SLAM pipeline we built upon, we include global bundle adjustment (BA) to address both the forgetting problem and camera drift during significant movements. We measure frame-to-frame motion using optical center distance $d$ and parallax angle $r$. When significant motion occurs, $d$ and $r$ increase dramatically. Therefore, keyframes are selected when these metrics exceed their respective thresholds $d_{l}$ and $r_l$.

\section{Experiments}

We now introduce our experiments and discuss the outcome. 

\textbf{Datasets}. We conduct our evaluation across three standard datasets: Replica \cite{straub2019replica}, ScanNet \cite{dai2017scannet}, and TUM RGB-D \cite{sturm2012benchmark}. Replica provides photo-realistic indoor scenes specifically designed for evaluating dense reconstruction quality. ScanNet offers extensively annotated real-world data across diverse indoor environments, ranging from small rooms to large scenes. The TUM RGB-D benchmark, instead, features small-scale indoor scenes captured with an external camera system, providing high-precision ground-truth trajectories.

\textbf{Baselines}. To support the generality of our approach, we first inject our DINO-informed design into different NeRF and GS-based SLAM systems characterized by various kinds of scene encoding, including grid-based \cite{wang2023co}, tri-planes \cite{johari2023eslam, zhu2024sni}, neural points \cite{sandstrom2023point}, 3D Gaussian Splats \cite{keetha2024splatam, matsuki2024gaussian}, and 2D Gaussian Surfels \cite{su2026gaus}. 
Furthermore, to fully assess the potential of DINO-SLAM, we compare it against several state-of-the-art methods, focusing on dense mapping, novel view synthesis, and camera tracking. 

\textbf{Metrics}. For quantitative evaluation, we adopt standard SLAM metrics \cite{tosi2024nerfs}: Absolute Trajectory Error (ATE) RMSE for tracking accuracy, and reconstruction quality metrics including \textit{accuracy (cm)}, \textit{completion (cm)}, \textit{completion rate (\%)}, \textit{depth l1 (cm)}. Novel view rendering quality is assessed using \textit{PSNR (dB)}, \textit{SSIM}, and \textit{LPIPS}.

\textbf{Implementation Details}. All experiments were conducted on a desktop PC equipped with an NVIDIA RTX 3090Ti GPU. In our NeRF-based pipeline, we employ a multi-resolution approach with geometry tri-planes at 24cm/6cm resolution and appearance/DINO tri-planes at 24cm/3cm resolution, the loss weights are configured as 5.0 for encoding feature loss, and 0.01 for DINO-feature loss, $k$ is set to 4 for keyframe selection. 
For our GS-based pipeline, we maintain consistency with the original Gaussian Splatting implementation \cite{kerbl20233d}, and set the DINO-feature loss weight to 1, $d_{l}$=0.7, and $r_l$=15.

\subsection{Ablation Experiments}

We conduct ablation studies to demonstrate the versatility of DINO-SLAM and assess the impact of each component in the framework. 

\textbf{DINO-SLAM Generality.} We prove the wide applicability of DINO-SLAM by integrating our SGE module with various neural implicit and explicit representations: hash-grids (a), tri-planes (b-c), neural points (d), 3D Gaussians Splats (e-f), and 2D Gaussian Surfels (g). 
Table \ref{tab:table4} shows consistent performance improvements on Replica across all encoding architectures when augmented with our SGE. 
Specifically, we can appreciate how our pipeline can further enhance the results achieved by SLAM systems already exploiting vanilla DINO features for semantic segmentation, such as SNI-SLAM (c), confirming how these features alone are insufficient and that lifting DINO with geometry awareness is crucial to achieve the largest gains. Furthermore, our design also significantly improves state-of-the-art systems such as GauS-SLAM (g), confirming the broad applicability of our SGE and suggesting the potential to extend our methodology to future SLAM systems based on either NeRF or GS. 
From now on, we will adopt ESLAM and GauS-SLAM to build our NeRF and GS variants of DINO-SLAM.

\begin{table}[t]
\renewcommand{\tabcolsep}{5pt}
\caption{\textbf{Ablation study -- DINO-SLAM generality.} We report some baselines and their DINO-extended variants, maintaining the same evaluation metrics reported in the original papers. * uses ground-truth semantic masks as supervision. }\label{tab:table4}\vspace{-0.3cm}
\resizebox{\columnwidth}{!}{
\begin{tabular}{clcccccc}
\hline
& Methods    & Encoding    & Acc.$\downarrow$  & Comp.$\downarrow$  & Comp.rate$\uparrow$  & Depth L1$\downarrow$      & RMSE(cm)$\downarrow$  \\ \hline
\multirow{2}{*}{(a)} & Co-SLAM \cite{wang2023co}    & Hash-grid   & 2.10           & 2.08  & 93.44     & 1.51          & 0.86          \\ 
& \textbf{+DINO} & Hash-grid   & \textbf{1.96}  & \textbf{2.06}  & \textbf{93.62} & \textbf{1.39} & \textbf{0.78} \\ \hdashline
\multirow{2}{*}{(b)} & ESLAM \cite{johari2023eslam}     & Tri-planes  & 2.18           & 1.75  & 96.46     & 0.94          & 0.63          \\
& \textbf{+DINO} & Tri-planes  & \textbf{1.88}  & \textbf{1.60}  & \textbf{97.22} & \textbf{0.63} & \textbf{0.49} \\ 
\hdashline
\multirow{2}{*}{(c)} & SNI-SLAM* \cite{zhu2024sni}  & Tri-planes  & 1.94        & 1.70  & 96.62     & 0.77          & 0.46          \\
& \textbf{+DINO} & Tri-planes  & \textbf{1.83}  & \textbf{1.61}  & \textbf{97.05} & \textbf{0.68} & \textbf{0.42} \\ 
\hline
& Methods    & Encoding    & F1$\uparrow$     & Pre.$\uparrow$  & Recall$\uparrow$    & Depth L1$\downarrow$      & RMSE(cm)$\downarrow$  \\ \hline
\multirow{2}{*}{(d)} & Point-SLAM \cite{sandstrom2023point} & Neur-points & 89.77          & 96.99 & 83.59     & 0.44          & 0.52          \\
& \textbf{+DINO} & Neur-points & \textbf{90.77} & \textbf{98.33} & \textbf{84.30}  & \textbf{0.41} & \textbf{0.46} \\ \hline
& Methods    & Encoding    & PSNR$\uparrow$   & SSIM$\uparrow$  & LPIPS$\downarrow$     & Depth L1$\downarrow$      & RMSE(cm)$\downarrow$  \\ \hline
\multirow{2}{*}{(e)} & SplaTAM \cite{keetha2024splatam}   & 3DGS        & 34.11          & 0.968  & 0.102       & 0.73          & 0.36          \\
 & \textbf{+DINO}      & 3DGS        & \textbf{34.65} & \textbf{0.971}  & \textbf{0.098}       & \textbf{0.66} & \textbf{0.33} \\ \hdashline
\multirow{2}{*}{(f)} & MonoGS \cite{matsuki2024gaussian}   & 3DGS        & 38.94          & 0.968  & 0.070       & 0.49       & 0.32          \\
& \textbf{+DINO}      & 3DGS        & \textbf{39.56} & \textbf{0.972}  & \textbf{0.047}    & \textbf{0.45}    & \textbf{0.29} \\ \hdashline
\multirow{2}{*}{(g)} & GauS-SLAM \cite{su2026gaus}   & 2DGS        & 40.25          & 0.991  & 0.027       & 0.43       & 0.06          \\
& \textbf{+DINO}      & 2DGS        & \textbf{41.42} & \textbf{0.994}  & \textbf{0.021}    & \textbf{0.12}    & \textbf{0.05} \\ 
\hline
\end{tabular}
}
\end{table}

\begin{table}[t]
\caption{\textbf{Ablation study -- Impact of DINO features.} We study the impact of features predicted by our SGE on the final accuracy on Replica. The NeRF-based pipelines (\textbf{upper}) evaluate reconstruction quality and camera tracking. The GS-based pipelines (\textbf{lower}) evaluate rendering quality and camera tracking. }\vspace{-0.3cm}
\renewcommand{\tabcolsep}{5pt}
\resizebox{\columnwidth}{!}{%
\begin{tabular}{clccccc}
\hline
& Component & Acc.$\downarrow$   & Comp.$\downarrow$   & Comp.rate$\uparrow$ & Depth L1$\downarrow$  & RMSE(cm)$\downarrow$ \\ \hline
(A) & ESLAM \cite{johari2023eslam}      & 2.18       & 1.75      & 96.46     & 0.94      & 0.63         \\
(B) & + vanilla DINO        & 2.07       & 1.67      & 96.63     & 0.79      & 0.58         \\
(B') & + concat(DINO,depth)  & 2.03 & 1.67  & 96.78      & 0.75     & 0.56 \\
(C) & \bf + geoDINO & \textbf{1.88}  & \textbf{1.60} & \textbf{97.22} & \textbf{0.63} & \textbf{0.49}    \\ \hline
& Component  & PSNR$\uparrow $     & SSIM$\uparrow $     & LPIPS$\downarrow$  & Depth L1$\downarrow$    & RMSE(cm)$\downarrow$ \\ \hline
(D) & GauS-SLAM \cite{su2026gaus}   & 40.25     & 0.991      & 0.027  & 0.43    & 0.06       \\
(E) & + vanilla DINO   & 40.78     & 0.992      & 0.025  & 0.27    & 0.06       \\
(F) & \bf + geoDINO      & \textbf{41.42}     & \textbf{0.994}      & \textbf{0.021}  & \textbf{0.12}    & \textbf{0.05}        \\
\hline
\end{tabular}
}
\label{tab:table5}
\end{table}

\textbf{Components Analysis.} To support our main claim on the importance of SGE to unveil the real potential of DINO features for SLAM, we conduct an ablation study to measure the impact of both vanilla DINO features and our enhanced geoDINO features, reporting the results achieved on Replica in Table \ref{tab:table5}. For the NeRF-based pipeline, we assess reconstruction and camera tracking performance, while for the GS-based pipeline, we evaluate the rendering quality and the accuracy of estimated poses.

\renewcommand{\tabcolsep}{10pt}
\begin{table*}[t]
\caption{\textbf{Tracking and mapping results on Replica.} We present comparative results for both our NeRF-based (\textbf{top}) and GS-based (\textbf{bottom}) implementations. For fair reconstruction evaluation, all generated meshes undergo post-processing using the culling strategy described in \cite{wang2023co}. Meshes marked with $\dag$ were obtained directly from the authors. The best results are highlighted as \colorbox[HTML]{B7D3B7}{\textbf{first}}, \colorbox[HTML]{D8E8C5}{second}, and \colorbox[HTML]{FFF3BB}{third}. }\vspace{-0.3cm}
\label{tab:table1}
    \centering
    \renewcommand{\tabcolsep}{5pt}
    \resizebox{\columnwidth}{!}{%
    \begin{tabular}{lccccc}
    \hline
    Method        & Acc.(cm)$\downarrow$ & Comp.(cm)$\downarrow$ & Comp.rate(\%)$\uparrow$ & Depth L1(cm)$\downarrow$ & RMSE(cm)$\downarrow$ \\ \hline
    iMAP \cite{sucar2021imap}        & 3.62     & 4.93  & 80.51     & 4.64          & 7.64         \\
    NICE-SLAM \cite{zhu2022nice}   & 2.37     & 2.65  & 91.14     & 1.90          & 2.50         \\
    Co-SLAM \cite{wang2023co}     & 2.10     & 2.08  & 93.44     & 1.51          & 0.86         \\
    ESLAM \cite{johari2023eslam}        & 2.18     & 1.75  & 96.46     & 0.94          & 0.63         \\
    Point-SLAM \cite{sandstrom2023point}   & \cellcolor[HTML]{B7D3B7}\textbf{1.41} & 3.10  & 88.89     & \cellcolor[HTML]{B7D3B7}\textbf{0.44} & \cellcolor[HTML]{D8E8C5}0.52         \\
    PLGSLAM$^{\dag}$ \cite{deng2024plgslam}     & 2.14     & \cellcolor[HTML]{FFF3BB}1.74  & 96.24     & 0.83  & 0.63   \\
    HS-SLAM \cite{hsslam}    & \cellcolor[HTML]{FFF3BB}1.96 & \cellcolor[HTML]{D8E8C5}1.67  & \cellcolor[HTML]{FFF3BB}96.57     & \cellcolor[HTML]{FFF3BB}0.71 & \cellcolor[HTML]{FFF3BB}0.53        \\
    Mamba-SLAM \cite{mamba-slam}   & 2.62 & 2.25  & 92.86     & - & -        \\  
    SNH-SLAM \cite{snh-slam}   & 2.08 & 1.78  & \cellcolor[HTML]{D8E8C5}96.85     & 1.15 & 0.61        \\
    QQ-SLAM \cite{qq-slam}   & 2.38 & 1.76  & 96.39     & 1.09 & 0.91        \\
    SNI-SLAM \cite{zhu2024sni}  & 1.94     & 1.70  & 96.62     & 0.77    & 0.46         \\
    
    \textbf{DINO-SLAM (NeRF)} & \cellcolor[HTML]{D8E8C5}1.88  & \cellcolor[HTML]{B7D3B7}\textbf{1.60} & \cellcolor[HTML]{B7D3B7}\textbf{97.22} & \cellcolor[HTML]{D8E8C5}0.63  & \cellcolor[HTML]{B7D3B7}\textbf{0.49}   \\ 
    
    \hline 
    \hline
    Method        & PSNR$\uparrow$    & SSIM$\uparrow$     & LPIP$\downarrow$   & Depth L1(cm)$\downarrow$        & RMSE(cm)$\downarrow$ \\ \hline
    SplaTAM \cite{keetha2024splatam}      & 34.11          & 0.968          & 0.102  & 0.73        & 0.36         \\
    Gaussian-SLAM \cite{yugay2023gaussian} & 38.90           & \cellcolor[HTML]{D8E8C5}0.993 & 0.069   & 0.68       & 0.31         \\
    MonoGS \cite{matsuki2024gaussian}        & 38.94          & 0.968          & 0.070    & 0.49       & 0.32         \\
    GS-SLAM \cite{yan2024gs}       & 34.27          & 0.975         & 0.082  & 1.16        & 0.50          \\  
    SEGS-SLAM \cite{segs-slam}       & \cellcolor[HTML]{FFF3BB}39.42          & 0.975         & \cellcolor[HTML]{B7D3B7}\textbf{0.021}    & -      & 0.43          \\ 
    GauS-SLAM \cite{su2026gaus}       & \cellcolor[HTML]{D8E8C5}40.25         & \cellcolor[HTML]{FFF3BB}0.991         & \cellcolor[HTML]{D8E8C5}0.027    & 0.43      & \cellcolor[HTML]{D8E8C5}0.06          \\     
    SemGauss \cite{zhu2024semgauss} & 35.03        & 0.982      & 0.062 & 0.50 & 0.33 \\     
    Hier-SLAM \cite{hier-slam}       & 35.70         & 0.980         & 0.067    & 0.49       & 0.32          \\ 
    VSS-SLAM \cite{vss-slam}       & 39.01         & 0.975         & 0.042     & \cellcolor[HTML]{D8E8C5}0.34      & 0.28          \\ 
    FGO-SLAM \cite{fgo-slam}       & 38.35         & 0.973         & 0.082    & 0.71      & -          \\
    FGS-SLAM \cite{fgs-slam}       & 38.75         & 0.974         & \cellcolor[HTML]{FFF3BB} 0.041    & -       & \cellcolor[HTML]{FFF3BB}0.15          \\
    CGS-SLAM \cite{cgs-slam}       & 34.44         & 0.980         & 0.090    & -       & 0.33         \\
    VSG-SLAM \cite{vsg-slam}       & 36.86         & 0.987         & 0.061    & \cellcolor[HTML]{FFF3BB}0.40      & 0.37          \\
    \textbf{DINO-SLAM (GS)}    & \cellcolor[HTML]{B7D3B7}\textbf{41.42} & \cellcolor[HTML]{B7D3B7}\textbf{0.994}          & \cellcolor[HTML]{B7D3B7}\textbf{0.021} & \cellcolor[HTML]{B7D3B7}\textbf{0.12}  & \cellcolor[HTML]{B7D3B7}\textbf{0.05} \\      
    \hline
    \end{tabular}
    }
\end{table*}

The results reveal a few key insights: by establishing ESLAM and GauS-SLAM without any DINO features as the baselines (A, D), we can observe some improvements obtained by naively embedding vanilla DINO features in both the NeRF and GS-based pipelines (B, E). These findings are not surprising and were already supported by existing literature on semantic SLAM \cite{zhu2024sni}. 
The lack of geometry awareness by DINO can be partially compensated by simply concatenating depth features $f_g$ (B'), although with limited effectiveness.
On the contrary, employing our SGE to obtain geoDINO features and using these latter significantly boosts the accuracy further, both for mapping/rendering and camera tracking, by doubling the improvement with respect to the one gained with vanilla DINO features on most metrics. This confirms our main claim and the importance of lifting DINO features with geometry awareness, something that cannot be achieved by simply concatenating depth features.

\begin{figure*}[t]
	\centering
	\includegraphics[width=\linewidth,scale=1.00]{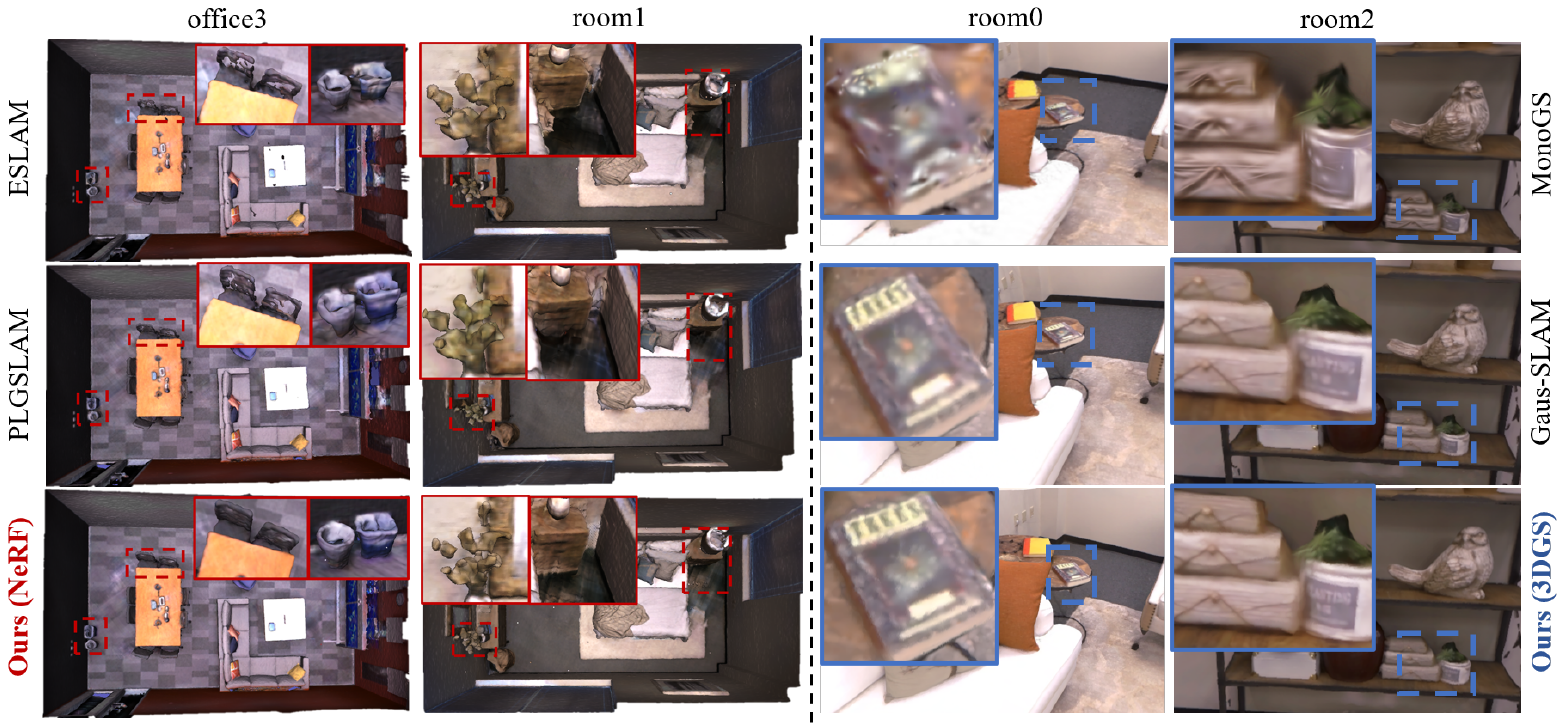}\vspace{-0.3cm}
	\caption{\textbf{Visualization of DINO-SLAM and baselines on the Replica dataset.} We present details of \textbf{reconstruction} and \textbf{rendering} quality with red boxes and blue boxes. The two \textbf{left columns} present the reconstruction performance of our \textbf{DINO-SLAM (NeRF)}. Our method yields superior mesh results with detailed textures and better completeness. 
    The two \textbf{right columns} show the rendering results of our \textbf{DINO-SLAM (GS)}. Our DINO-SLAM demonstrates high-fidelity and realistic images, especially on text tags and object textures.
    }
	\label{fig:vis4comparison}
\end{figure*}

\subsection{Comparison with State-Of-The-Art}

We now compare our NeRF and GS variants of DINO-SLAM, built respectively on top of ESLAM and GauS-SLAM, against state-of-the-art SLAM frameworks. 

\textbf{Results on Replica.}
We begin our experiments on Replica, running our pipelines on 8 diverse scenes following standard practice in the literature. 
Table \ref{tab:table1} compares our DINO-SLAM variants against state-of-the-art SLAM approaches. The table is divided into two sections: at the top, it presents reconstruction and tracking results for NeRF-based methods, while at the bottom, it shows rendering and tracking performance for GS-based methods. All reported values represent averages across the test scenes. 
The results highlight the effectiveness of our DINO-SLAM paradigms. In particular, our GS-based variant achieves state-of-the-art performance in both rendering quality and tracking accuracy among all GS-based methods. Similarly, our NeRF-based implementation excels in completion, completion rate, and RMSE metrics, while remaining competitive in accuracy and depth L1 error. 
Figure \ref{fig:vis4comparison} collects qualitative results that further validate the superiority of DINO-SLAM. The meshes produced by our NeRF-based variant (the two left columns) demonstrate superior surface detail preservation while successfully completing previously missing regions. The rendering results of our GS-based variant (the two right columns) show photo-realistic rendering quality with significantly improved texture detail resolution.

\textbf{Robustness to depth noise.} Since geoDINO enhances DINO with structural knowledge derived from depth, the presence of noise on this latter can harm our framework.
In Tab. \ref{tab:table_noise}, we compare the performance achieved by GauS-SLAM and DINO-SLAM (GS) when processing depth maps corrupted with Gaussian noise ($\mu=0,\sigma=0.1$). Despite the accuracy drops for both, our DINO-SLAM processing noisy depth is still better than GauS-SLAM with noise-free depth, demonstrating that our SGE enjoys some robustness to depth noise.

\textbf{Results on ScanNet.}
Our evaluation extends to ScanNet, where we test both DINO-SLAM variants on six sequences of varying complexity, including challenging long trajectories exceeding 5000 frames. 
As shown in Table \ref{tab:table2}, our GS-based pipeline significantly outperforms existing GS-based systems, while our NeRF-based pipeline still surpasses most baselines. In this sense, we note that PLGSLAM achieves slightly better results due to its design being specifically tailored for larger scene reconstruction, whereas our contribution is orthogonal to it and could potentially be  integrated with such a framework as well.

\begin{table}[t]
\caption{\textbf{Robustness to depth noises.} Results on Replica \textit{office1}.}\vspace{-0.3cm}
\label{tab:table_noise}
\resizebox{\columnwidth}{!}{
\begin{tabular}{cccccc}
    \hline
               & PSNR $\uparrow$ & SSIM $\uparrow$  & LPIPS $\downarrow$ & Depth L1 $\downarrow$ & RMSE $\downarrow$ \\ \hline
    Gaus-SLAM      & 41.40 & 0.981 & 0.055 & 0.37     & 0.04 \\
    w/ Depth Noise & 39.87 & 0.977 & 0.062 & 0.41     & 0.05 \\
    \hline
    DINO-SLAM (GS) & 44.65 & 0.994 & 0.023 & 0.22     & 0.02 \\
    w/ Depth Noise & 42.38 & 0.986 & 0.029 & 0.29     & 0.03 \\ \hline
    \end{tabular}} 
\end{table}

\begin{table}[t]
\caption{\textbf{Tracking results on ScanNet.} Avg. denotes the average ATE RMSE. }\vspace{-0.3cm}
\label{tab:table2}
\resizebox{\columnwidth}{!}{%
\renewcommand{\tabcolsep}{15pt}
\begin{tabular}{lccccccc}
\hline
Method          & 0000         & 0059         & 0106         & 0169         & 0181          & 0207         & Avg.          \\ \hline
iMAP \cite{sucar2021imap}           & 55.9         & 32.0         & 17.5         & 70.5         & 32.1          & 11.9         & 36.7          \\
NICE-SLAM  \cite{zhu2022nice}     & 12.0           & 14.0           & 7.9          & 10.9         & 13.4          & 6.2          & 10.7          \\
Co-SLAM  \cite{wang2023co}       & \cellcolor[HTML]{FFF3BB}7.1        & 11.1         & 9.4          & \cellcolor[HTML]{B7D3B7}\textbf{5.9}    & 11.8          & 7.1          & 8.7           \\
ESLAM  \cite{johari2023eslam}       & 7.3          & 8.5          & \cellcolor[HTML]{D8E8C5}7.5          & \cellcolor[HTML]{D8E8C5}6.5          & \cellcolor[HTML]{D8E8C5}9.0           & \cellcolor[HTML]{D8E8C5}5.7          & \cellcolor[HTML]{FFF3BB}7.4           \\
PLGSLAM$\dag$ \cite{deng2024plgslam}   & -    & -     & -    & -   & -  & - & \cellcolor[HTML]{B7D3B7}\textbf{6.8}  \\
Point-SLAM \cite{sandstrom2023point}   & 10.2         & \cellcolor[HTML]{D8E8C5}7.8          & 8.7          & 22.2         & 14.8          & 9.5          & 12.2           \\
Mamba-SLAM \cite{mamba-slam}    & -        & 10.0         & 7.8          & 10.1         & 12.4          & \cellcolor[HTML]{B7D3B7} \textbf{4.7}          & -           \\
SNH-SLAM \cite{snh-slam}   & \cellcolor[HTML]{B7D3B7} \textbf{6.7}         & \cellcolor[HTML]{FFF3BB}8.1         & \cellcolor[HTML]{FFF3BB}7.7         & \cellcolor[HTML]{FFF3BB}7.3         & \cellcolor[HTML]{FFF3BB} 10.6          & 6.3          & 7.8           \\
QQ-SLAM \cite{qq-slam}   & \cellcolor[HTML]{D8E8C5}7.0         & 9.5          & 8.8         & \cellcolor[HTML]{D8E8C5}6.5         & 13.3          & \cellcolor[HTML]{FFF3BB}5.9          & 8.5           \\
\textbf{DINO-SLAM (NeRF)} & 7.6          & \cellcolor[HTML]{B7D3B7}\textbf{7.5}          & \cellcolor[HTML]{B7D3B7}\textbf{7.0}          & \cellcolor[HTML]{B7D3B7}\textbf{5.9}          & \cellcolor[HTML]{B7D3B7}\textbf{8.6}  & \cellcolor[HTML]{B7D3B7}\textbf{4.7} & \cellcolor[HTML]{D8E8C5}6.9           \\ 
\hline \hline
SplaTAM  \cite{keetha2024splatam}            & 12.8         & 10.1         & 17.7         & 12.1         & 11.1 & 7.5          & 11.9          \\
MonoGS   \cite{matsuki2024gaussian}            & \cellcolor[HTML]{FFF3BB}9.8 & 32.1         & \cellcolor[HTML]{FFF3BB}8.9          & \cellcolor[HTML]{FFF3BB}10.7         & 21.8          & 7.9 & 15.2          \\
GauS-SLAM  \cite{su2026gaus}      & \cellcolor[HTML]{D8E8C5}9.3         & \cellcolor[HTML]{D8E8C5}7.1       & \cellcolor[HTML]{D8E8C5}7.0         & \cellcolor[HTML]{D8E8C5}7.4         & 17.5            & \cellcolor[HTML]{FFF3BB}6.2         & \cellcolor[HTML]{D8E8C5}9.1          \\
Hier-SLAM \cite{hier-slam}      & 11.5         & 9.6       & 17.8         & 11.9         & \cellcolor[HTML]{B7D3B7} \textbf{10.0}            & 7.3         & 11.4          \\
CGS-SLAM \cite{cgs-slam}      & 11.3         & \cellcolor[HTML]{FFF3BB}9.2       & 16.3         & 11.0         & 10.8            & 6.5         & \cellcolor[HTML]{FFF3BB}10.8          \\
VSG-SLAM \cite{vsg-slam}      & 11.6         & -       & -         & 11.0         & \cellcolor[HTML]{FFF3BB}10.7            & \cellcolor[HTML]{D8E8C5}6.1         & -          \\

\textbf{DINO-SLAM (GS)}       & \cellcolor[HTML]{B7D3B7}\textbf{8.5}        & \cellcolor[HTML]{B7D3B7}\textbf{6.5} & \cellcolor[HTML]{B7D3B7}\textbf{6.2} & \cellcolor[HTML]{B7D3B7}\textbf{6.6} & \cellcolor[HTML]{D8E8C5}10.5          & \cellcolor[HTML]{B7D3B7}\textbf{5.8}       & \cellcolor[HTML]{B7D3B7}\textbf{7.3} \\
\hline
\end{tabular}
}
\end{table}

\begin{table}[t]
\caption{\textbf{Tracking results on TUM.} Avg. denotes the average ATE RMSE. }
\label{tab:table3}\vspace{-0.3cm}
\renewcommand{\tabcolsep}{20pt}
\resizebox{\columnwidth}{!}{%
\begin{tabular}{lcccc}
\hline
Method          & fr1\_desk      & fr2\_xyz       & fr3\_office     & Avg.          \\ \hline
iMAP \cite{sucar2021imap}                 & 4.90          & 2.00             & 5.80           & 4.23          \\
NICE-SLAM \cite{zhu2022nice}     & 2.70          & 1.80           & 3.00             & 2.50           \\
Co-SLAM \cite{wang2023co}        & \cellcolor[HTML]{D8E8C5}2.40          & 1.70          & \cellcolor[HTML]{D8E8C5}2.40          & \cellcolor[HTML]{FFF3BB}2.17          \\
ESLAM \cite{johari2023eslam}        & \cellcolor[HTML]{FFF3BB}2.47     & \cellcolor[HTML]{B7D3B7}\textbf{1.11}        & \cellcolor[HTML]{FFF3BB}2.42        & \cellcolor[HTML]{D8E8C5}2.00          \\
Point-SLAM \cite{sandstrom2023point}   & 4.34          & \cellcolor[HTML]{FFF3BB}1.31          & 3.48          & 3.04          \\
QQ-SLAM \cite{qq-slam}   & 2.61          & 1.70          & 2.70          & 2.34          \\
\textbf{DINO-SLAM (NeRF)} & \cellcolor[HTML]{B7D3B7}\textbf{2.39}  & \cellcolor[HTML]{D8E8C5}1.14  & \cellcolor[HTML]{B7D3B7}\textbf{2.27}     & \cellcolor[HTML]{B7D3B7}\textbf{1.93}  \\
\hline \hline
SplaTAM  \cite{keetha2024splatam}           & 3.35           & \cellcolor[HTML]{FFF3BB}1.24           & 5.16           & 3.27          \\
MonoGS  \cite{matsuki2024gaussian}            & \cellcolor[HTML]{B7D3B7}\textbf{1.50}          & 1.44     & 1.49      & \cellcolor[HTML]{D8E8C5}1.47    
\\
GS-SLAM  \cite{yan2024gs}           & 3.30           & 1.30           & 6.60           & 3.73          \\
SEGS-SLAM \cite{segs-slam}            & 3.19           & \cellcolor[HTML]{B7D3B7}\textbf{0.37}           & \cellcolor[HTML]{B7D3B7}\textbf{1.03}           & 1.53          \\
GauS-SLAM \cite{su2026gaus}            & 1.82           & 1.34           & \cellcolor[HTML]{FFF3BB}1.46           & 1.54          \\
FGS-SLAM  \cite{fgs-slam}           & 2.50          & 1.50           & 2.10           & 2.00          \\
VSS-SLAM  \cite{vss-slam}           & \cellcolor[HTML]{D8E8C5}1.56          & 1.39           & 1.51           & \cellcolor[HTML]{FFF3BB}1.49          \\
\textbf{DINO-SLAM (GS)}   & \cellcolor[HTML]{FFF3BB}1.77  & \cellcolor[HTML]{D8E8C5}1.07 & \cellcolor[HTML]{D8E8C5}1.41      & \cellcolor[HTML]{B7D3B7}\textbf{1.42} \\
\hline
\end{tabular}%
}
\end{table}

\textbf{Results on TUM.}
To further validate our approach, we run experiments on three widely-used sequences from the TUM RGB-D dataset, including both short and long trajectories. As shown in Table \ref{tab:table3}, although not excelling on all of the single sequences, both implementations of DINO-SLAM achieve superior pose estimation accuracy on average, outperforming all existing methods. These results confirm that our DINO-SLAM pipelines excel across all standard benchmarks commonly adopted in the field.

\begin{table}[t]

\caption{\textbf{Memory usage and runtime} on Replica \textit{room0}.}
\label{tab:performance}\vspace{-0.3cm}
\renewcommand{\tabcolsep}{15pt}
\resizebox{\columnwidth}{!}{
\begin{tabular}{lcccc}
\hline
Method       & Size(MB)$\downarrow$  & FPS$\uparrow$ & Comp.$\downarrow$ & RMSE$\downarrow$     \\ \hline
Co-SLAM \cite{wang2023co}         & \textbf{6.7}          & \textbf{7.97} & 2.08 & 0.86 \\
Point-SLAM  \cite{sandstrom2023point}  & 54.8          & 0.23  & 3.10 & 0.52 \\ 
ESLAM    \cite{johari2023eslam}      & 27.2         & 4.62   & 1.75 & 0.63\\
\textbf{DINO-SLAM (NeRF)}  & 48.7          & 2.55 & \textbf{1.60} & \textbf{0.49} \\ \hline\hline
Method       & Size(MB)$\downarrow$  & FPS$\uparrow$ & PSNR$\uparrow$ & RMSE$\downarrow$     \\ \hline 
MonoGS  \cite{matsuki2024gaussian}               & \textbf{28.5}          & \textbf{0.62}  & 38.94 & 0.32 \\
SplaTAM  \cite{keetha2024splatam}            & 265.3         & 0.14   & 34.11   & 0.36\\
GauS-SLAM  \cite{su2026gaus}            & 347.0      & 0.59   & 40.25   & 0.06\\
\textbf{DINO-SLAM (GS)}    & 343.1          & 0.58    & \textbf{41.42} & \textbf{0.05}\\
\hline
\end{tabular}}
\end{table}

\subsection{Performance Analysis}
We compare our DINO-SLAM pipelines with NeRF-based and GS-based baselines in both FPS (average time required to process per frame in a sequence) and memory usage (total footprint of the encoder and decoder). As shown in Table \ref{tab:performance}, Co-SLAM is the fastest method, yet its reconstruction accuracy is significantly lower than our NeRF-based SLAM. In contrast, our DINO-SLAM (NeRF) achieves better results with moderate memory requirements, although halving the FPS with respect to its baseline, ESLAM. 
Besides, concerning GS methods, we can appreciate how DINO-SLAM achieves the best results, with negligible overhead on top of the GauS-SLAM baseline with respect to the complexity of the GS engines.

We conclude with a detailed computational breakdown in Table \ref{tab:breakthrough}, to precisely quantify the overhead introduced by our SGE and the overall DINO-SLAM paradigm. At the top (a), we report the FPS achieved by the different SLAM baselines and their DINO-SLAM counterparts, highlighting a negligible difference in most cases. In the middle (b), we present a detailed runtime analysis of our DINO-SLAM (GS) framework, quantifying the moderate overhead introduced by geoDINO features and our SGE module throughout the optimization pipeline. Finally, at the bottom (c), we report the memory footprint of DINO-SLAM (GS) on different scenes in ScanNet. Even in large scenes, the memory footprint is acceptable, thanks to the pruning strategy already employed by GauS-SLAM.

\begin{table}[t]
\caption{\textbf{Computational breakdown.} (a) {Overhead introduced by SGE.} FPS is reported on Replica \textit{room0}. (b)
{Runtime breakdown (s) for DINO-SLAM (GS)} on Replica \textit{office1}. (c) {Memory footprint of DINO-SLAM (GS)} on ScanNet.}\label{tab:breakthrough}\vspace{-0.3cm}
\renewcommand{\tabcolsep}{5pt}
\resizebox{\columnwidth}{!}{
    \begin{tabular}{cc}
    \multirow{3}{*}{\textbf{(a)}} &
    \begin{tabular}{cccccccc}
    \hline
               & Co-SLAM & ESLAM & SNI-SLAM & Point-SLAM & SplaTAM & MonoGS & GauS-SLAM \\ \hline
    w/o DINO   & 7.97    & 4.62  & 0.61     & 0.23       & 0.14      & 0.62   & 0.59      \\
    w/ geoDINO & 7.01    & 2.55  & 0.56     & 0.20       & 0.12      & 0.61   & 0.58      \\ \hline
    \end{tabular} \\ \\
    \multirow{3}{*}{\textbf{(b)}} &
    \renewcommand{\tabcolsep}{10pt}
    \begin{tabular}{cccccc}
    \hline
           & SGE Forward & Backward & Rendering & Per Frame & Whole Sequence \\ \hline
    w/o DINO (s)   & -           & 0.004            & 0.021     & 1.69 & 3387     \\
    w/ geoDINO (s) & 0.041      & 0.007        & 0.033     & 1.75   & 3510    \\ \hline
    \end{tabular} \\ \\

    \multirow{2}{*}{\textbf{(c)}} &
    \renewcommand{\tabcolsep}{17pt}
    \begin{tabular}{ccccccc}
    \hline
        & 0000  & 0059  & 0106  & 0169  & 0181  & 0207  \\ \hline
    Memory (MB) & 356.7 & 216.2 & 225.1 & 173.6 & 292.3 & 317.5 \\ \hline
    \end{tabular}
    
    \end{tabular}} 
\end{table}
\section{Conclusion} We presented DINO-SLAM, a DINO-informed dense RGB-D SLAM design paradigm that enhances both neural implicit and explicit representations. 
We implemented two foundational paradigms suited for both NeRF-based and GS-based SLAM systems built on top of our SGE module, which injects vanilla DINO features with geometry awareness to better capture structural relationships.
Our experiments validate the wide applicability and the superior performance achieved by our method. 
In future work, we will extend DINO-SLAM with more advanced components orthogonal to its design, such as loop-closure and sub-mapping.


\bibliographystyle{splncs04}
\bibliography{main}
\end{document}